\documentclass[11pt,a4paper]{article}
\usepackage[hyperref]{acl2017_arxiv}
\usepackage{times}
\usepackage{svg}
\usepackage{url}
\usepackage{latexsym}
\usepackage{graphicx}
\usepackage{hyperref}
\usepackage{enumitem}
\usepackage{sidecap}
\usepackage{amsmath}
\usepackage[small,compact]{titlesec}
\usepackage[hang,flushmargin]{footmisc}
\setenumerate{label=(\roman*),itemsep=1pt,topsep=1pt}
\usepackage{lipsum}

  \renewenvironment{thebibliography}[1]{%
    \begin{oldthebibliography}{#1}%
      \setlength{\parskip}{0ex}%
      \setlength{\itemsep}{0ex}%
  }%
  {%
    \end{oldthebibliography}%
  }
\title{Scattertext: a Browser-Based Tool for Visualizing how Corpora Differ}
\author{Jason S. Kessler \\
  CDK Global \\
  {\tt jason.kessler@gmail.com}  \\}
\date{}
\begin{document}
\maketitle
\begin{abstract}
Scattertext is an open source tool for visualizing linguistic variation between document categories in a language-independent way. The tool presents a scatterplot, where each axis corresponds to the rank-frequency a term occurs in a category of documents.  Through a tie-breaking strategy, the tool is able to display thousands of visible term-representing points and find space to legibly label hundreds of them.   Scattertext also lends itself to a query-based visualization of how the use of terms with similar embeddings differs between document categories, as well as a visualization for comparing the importance scores of bag-of-words features to univariate metrics.
\end{abstract}
\section{Introduction}
Finding words and phrases that discriminate categories of text is a common application of statistical NLP. For example, finding words that are most characteristic of a political party in congressional speeches can help political scientists identify means of partisan framing \cite{monroe08,grimmer2010}, while identifying differences in word usage between male and female characters in films can highlight narrative archetypes \cite{schofield2016gender}.  Language use in social media can inform understanding of personality types \cite{Schwartz13}, and provides insights into customers' evaluations of restaurants” \cite{jurafsky2014}.

A wide range of visualizations have been used to highlight discriminating words-- simple ranked lists of words, word clouds, word bubbles, and word-based scatter plots.  These techniques have a number of limitations.  For example, the difficulty in comparing the relative frequencies of two terms in a word cloud, or in legibly displaying term labels in scatterplots.
\begin{figure*}[h!!]
  \includegraphics[width=\linewidth,scale=0.8]{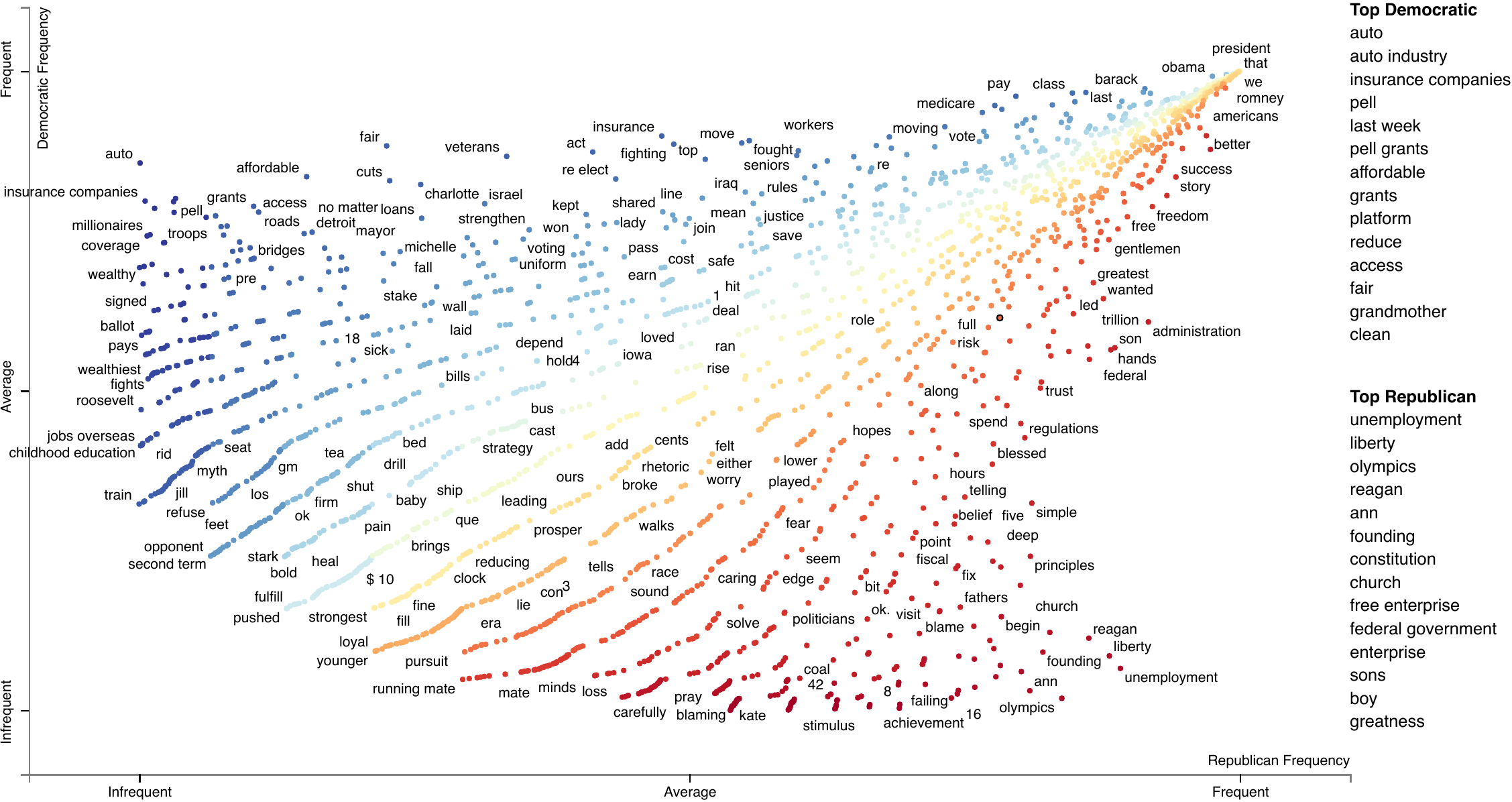}
  \caption{Scattertext visualization of words and phrases used in the 2012 Political Conventions.  2,202 points are colored red or blue based on the association of their corresponding terms with Democrats or Republicans, 215 of which were labeled. The corpus consists of 123 speeches by Democrats (76,864 words) and 66 by Republicans (58,138 words). The most associated terms are listed under ``Top Democrat'' and ``Top Republican'' headings. Interactive version: \href{https://jasonkessler.github.io/st-main.html}{https://jasonkessler.github.io/st-main.html}}
\label{scattertextmain}\vspace{-5mm}
\end{figure*}

Scattertext\footnote{\href{http://www.github.com/JasonKessler/scattertext}{github.com/JasonKessler/scattertext}} is an interactive, scalable tool which overcomes many of these limitations.  It is built around a scatterplot which displays a high number of words and phrases used in a corpus.  Points representing terms are positioned to allow a high number of unobstructed labels and to indicate category association.  The coordinates of a point indicate how frequently the word is used in each category.  

Figure \ref{scattertextmain} shows an example of a Scattertext plot comparing Republican and Democratic political speeches.  The higher up a point is on the y-axis, the more it was used by Democrats, and similarly, the further right on the x-axis a point appears, the more its corresponding word was used by Republicans.  Highly associated terms fall closer to the upper left and lower right-hand corners of the chart, while stop words fall in the far upper right-hand corner.  Words occurring infrequently in both classes fall closer to the lower left-hand corner.  
When used interactively, mousing-over a point shows statistics about a term's relative use in the two contrasting categories, and clicking on a term shows excerpts from convention speeches used.  

The point placement, intelligent word-labeling, and auxiliary term-lists ensure a low-whitespace, legible plot.  These are issues which have plagued other scatterplot visualizations showing discriminative language.

\S\ref{onviz} discusses different views of term-category association that make up the basis of visualizations.  In \S\ref{pastwork}, the objectives, strengths, and weaknesses of existing visualization techniques. \S\ref{scattertext} presents the technical details behind Scattertext. \S\ref{embeddings} discusses how Scattertext can be used to identify category-discriminating terms that are semantically similar to a query.
\section{On text visualization}
\label{onviz}
The simplest visualization, a list of words ranked by their scores, is easy to produce, interpret and is thus very common in the literature.  There are numerous ways of producing word scores for ranking which are thoroughly covered in previous work.  The reader is directed to Monroe et al. \shortcite{monroe08} (subsequently referred to as MCQ) for an overview of model-based term scoring algorithms.  Also of interest, Bitvai and Cohn \shortcite{Bitvai15} present a method for finding sparse words and phrase scores from a trained ANN (with bag-of-words features) and its training data. 

Regardless of how complex the calculation, word scores capture a number of different measures of word-association, which can be interesting when viewed independently instead of as part of a unitary score.  These loosely defined measures include:  \vspace{-0.1in}

\begin{description}[style=unboxed,leftmargin=0cm]
\item[Precision] A word's discriminative power regardless of its frequency.  A term that appears once in the categorized corpus will have perfect precision. This (and subsequent metrics) presuppose a balanced class distribution.  Words close to the x and y-axis in Scattertext have high precision.  \vspace{-0.1in}
\item[Recall] The frequency a word appears in a particular class, or $P(\mbox{word}|\mbox{class})$.  The variance of precision tends to decrease as recall increases.  Extremely high recall words tend to be stop-words.  High recall words occur close to the top and right sides of Scattertext plots.  \vspace{-0.1in}
\item[Non-redundancy] The level of a word's discriminative power given other words that co-occur with it.  If a word $w_a$ always co-occurs with $w_b$ and word $w_b$ has a higher precision and recall, $w_a$ would have a high level of redundancy. Measuring redundancy is non-trivial, and has traditionally been approached through penalized logistic regression \cite{joshi2010}, as well as through other feature selection techniques.  In configurations of Scattertext such as the one discussed at the end of \S\ref{scattertext}, terms can be colored based on their regression coefficients that indicate non-redundancy.  \vspace{-0.1in}
\item[Characteristicness] How much more does a word occur in than the categories examined than in background in-domain text?  For example, if comparing positive and negative reviews of a single movie, a logical background corpus may be reviews of other movies. Highly associated terms tend to be characteristic because they frequently appear in one category and not the other. Some visualizations explicitly highlight these, ex. \cite{vennclouds}. \vspace{-0.1in}
\end{description}
\section{Past work and design motivation}
\label{pastwork}
Text visualizations manipulate the position and appearance of words or points representing them to indicate their relative scores in these measures. For example, in Schwartz et al. \shortcite{Schwartz13}, two word clouds are given, one per each category of text being compared.  Words (and selected n-grams) are sized by their linear regression coefficients (a composite metric of precision, recall, and redundancy) and colored by frequency. Only words occurring in $\geq$1\% of documents and having Bonferroni-corrected coefficient p-values of $<$0.001 were shown. Given that these words are highly correlated to their class of interest, the frequency of use is likely a good proxy for recall.

Coppersmith and Kelly \shortcite{vennclouds} also describe a word-cloud based visualization for discriminating terms, but intend it for categories which are both small subsets of a much larger corpus. They include a third, middle cloud for terms that appear characteristic.  

Word clouds can be difficult to interpret.  It is difficult to compare the sizes of two non-horizontally adjacent words,  as well as the relative color intensities of any two words. Longer words unintentionally appear more important since they naturally occupy more space in the cloud.  Sizing of words can be a source of confusion when used to represent precision, since a larger word may naturally be seen as more frequent.

Bostock et al. \shortcite{Bostock2012}\footnote{\href{http://www.nytimes.com/interactive/2012/09/06/us/politics/convention-word-counts.html}{nytimes.com/interactive/2012/09/06/us/politics/convention-word-counts.html}} features an interactive word-bubble visualization for exploring different word usage among Republicans and Democrats in the 2012 US presidential nominating conventions.  Each term displayed is represented by a bubble, sized proportionate to their frequency.  Each bubble is colored blue and red, s.t. the blue partition's size corresponds to the term's relative use by Democrats.  Terms were manually chosen, and arranged along the x-axis based on their discriminative power. When clicked, sentences from speeches containing the word used are listed below the visualization. 

The dataset used in Bostock et al. \shortcite{Bostock2012} is used to demonstrate the capabilities of Scattertext in each of these figures.  The dataset is available via the Scattertext Github page.
\subsection{Scatterplot visualizations}
\vspace{-4mm}
\begin{figure}[h] 
\includegraphics[width=\columnwidth]{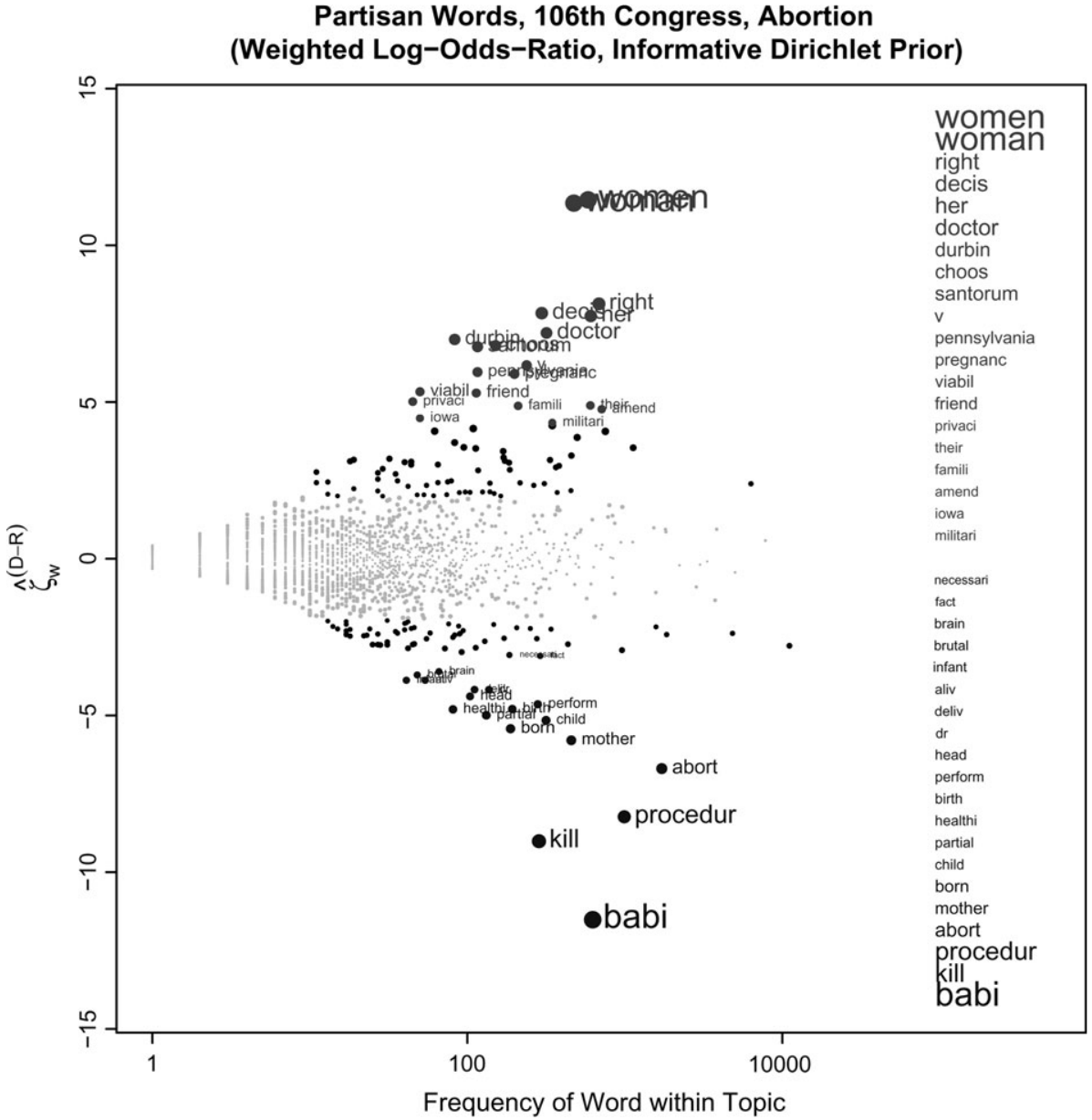}
\includegraphics[width=\columnwidth]{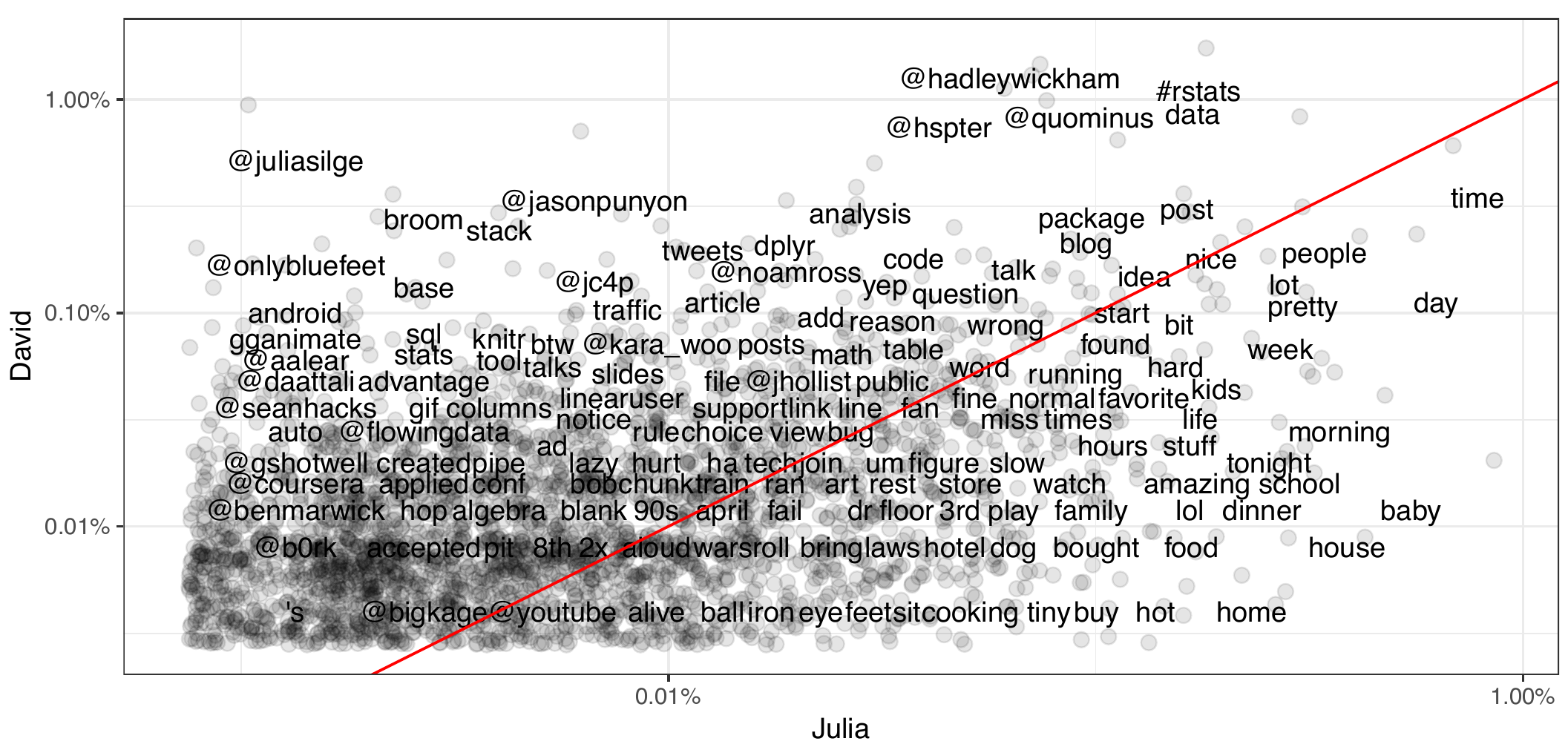}
\caption{A sample of existing scatterplot visualizations. MCQ's is at the top. Tidytext is below.} 
\label{scatters}
\end{figure}
\vspace{-3mm}
MCQ present a visualization to illustrate the use of their proposed word score, log-odds-ratio with an informative Dirichlet prior (top of Figure \ref{scatters}).  This visualization plots word-representing points along two axes.  The axes are log$_{10}$ recall vs. the difference in word scores z-scores.  Points with a z-score difference $<$1.96 are grayed-out, while the top and bottom 20 are labeled, both by each point and on the right-hand side.  The side-labeling is necessary because labels are permitted to overlap, hindering their on-plot readability.  The sizes of points and labels are increased proportionally to the word score. This word score encompasses precision, recall, and characteristicness since it penalizes scores of terms used more frequently in the background corpus. MCQ used this type of plot to illustrate the different effects of various scoring techniques introduced in the paper.  However, the small number of points which are possible to label limit its utility for in-depth corpus analysis.  

Schofield and Mehr \shortcite{schofield2016gender} use essentially the same visualization, but plot over 100 corresponding n-grams next to an unlabeled frequency/z-score plot.  While this is appropriate for publication, displaying associated terms and the shape of the score distribution, it is impossible to align all but the highest scoring points to their labels. 

The tidytext R-package \cite{tidytext} documentation includes a non-interactive ggplot2-based scatter plot that is very similar to Scattertext.   The x and y-axes both, like in Scattertext, correspond to word frequencies in the two contrasting categories, with jitter added.\footnote{This type of visualization may have first been introduced in Rudder \shortcite{Rudder2014}.}  In the example in Figure \ref{scatters} (bottom), the contrasting categories are tweets from two different accounts.  The red diagonal line separates words based on their odds-ratio.  Importantly, compared to MCQ, less of this chart's area is occupied by whitespace. 

While tidytext's labels do not overlap each other (in contrast to MCQ) they do overlap points.  The points' semi-transparency makes labels in less-dense areas legible, the dense interior of the chart is nearly illegible, with both points and labels obscured. 
\begin{figure}[h]
\vspace{-.25cm}
\includegraphics[width=\columnwidth]{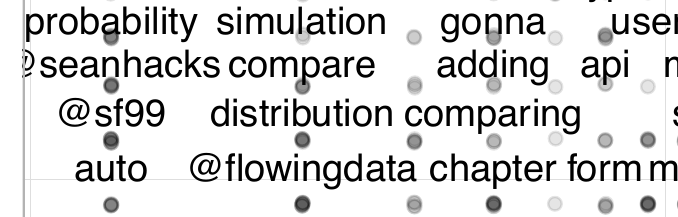}
\caption{A small cropping from an un-jittered version at the bottom of Figure \ref{scatters}.  The dark, opaque points indicate stacks of points.}
\label{nojitterfig}
\vspace{-.25cm}
\end{figure}
Figure \ref{nojitterfig} shows an excerpt of the same plot, but with no jitter.  Words appearing with the same frequency in both categories all become stacked atop each other, however, this provides more interior space for labeling. 

As a side note, many text visualizations plot words in a 2D space according to their similarity in a high dimensional space.  For example, Cho et al. \citeyear{Cho2014} uses the Barnes-Hut-SNE to plot words in a 2D space s.t. those with similar representations are grouped close together. Class-association does not play a role in this line of research, and global position is essentially irrelevant. 

The next section presents Scattertext and how its approach to word ordering solves the problems discussed above. 
\section{Scattertext}
\label{scattertext}
Scattertext builds on tinytext and Rudder \shortcite{Rudder2014}.  It plots a set of unigrams and bigrams (referred to in this paper as ``terms'') found in a corpus of documents assigned to one of two categories on a two-dimensional scatterplot.  

In the following notation, user-supplied parameters are in bold typeface. 

Consider a corpus of documents $C$ with disjoint subsets $A$ and $B$ s.t. $A \cup B \equiv C$. Let $\phi^T(t,C)$ be the number of times term $t$ occurs in $C$, $\phi^{T}(t,A)$ be the the number of times $t$ occurs in $A$. Let $\phi^{D}(t,A)$ refer to the number of documents in $A$ containing $t$. Let $t_{ij}$ be the $j$th word in term $t_i$. In practice, $j \in \{1,2\}$. The parameter $\mbox{\boldmath$\phi$}$ may be $\phi^T$ or $\phi^D$.\footnote{$\phi^D$ is useful when documents contain unique, characteristic, highly frequent terms.  For example, names of movies can have high $\phi^T$ when finding differences in positive and negative film reviews. The may lead to them receiving higher scores than sentiment terms.}  Other feature representations (ex., tf.idf) may be used for $\mbox{\boldmath$\phi$}$. 
\begin{equation}
\vspace{-.2cm}
Pr[t_i] = \frac{\mbox{\boldmath$\phi$}(t_i,C)}{\sum_{t \in C\wedge|t|\equiv|t_i|} \mbox{\boldmath$\phi$}(t,C)}.
\vspace{-.1cm}
\end{equation}
The construction of the set of terms included in the visualization $V$ is a two-step process. Terms must occur $geq$\mbox{\boldmath$m$} times, and if bigrams, appear to be phrases.  In order to keep the approach language neutral, I follow Schartz et al. \shortcite{Schwartz13}, and use a pointwise mutual information score to filter out bigrams that do not occur far more frequently than would be expected.  Let
\begin{equation}
\vspace{-.2cm}
PMI(t_i) = \log \frac{Pr[t_i]}{\prod_{t_{ij}\in t_i{Pr[T_{ij}]}}}.
\end{equation}
The minimum $PMI$ accepted is \mbox{\boldmath$p$}. Now, $V$ can be defined as 
\begin{equation}
\{t|\mbox{\boldmath$\phi$}(t,C)\geq\mbox{\boldmath$m$} \wedge (|t| \equiv 1 \vee PMI(t) > \mbox{\boldmath$p$})\}
\end{equation}
Let a term $t$'s coordinates on the scatterplot be $(x^{A}_{t}, x^{B}_t)$, where $A$ and $B$ are the two document categories. Although $x^{K}_t$ is proportional to $\mbox{\boldmath$\phi$}(t,K)$, many terms will have identical $\mbox{\boldmath$\phi$}(t,K)$ values.  To break ties the word that appears last alphabetically will have a larger $x^{K}_t$.

Let us define $r^{K}_t$ s.t. $t \in V$ and $K \in \{A,B\}$ as the ranks of $\mbox{\boldmath$\phi$}(t,K)$, sorted in ascending order, where ties are broken by terms' alphabetical order.  This allows us to define \vspace{-.2cm}
\begin{equation}
\vspace{-.2cm}
x^{K}_t = \frac{r^{K}_t}{\arg\!\max r^{K}}
\end{equation}
This limits $x$ values to $[0,1]$, ensuring both axes are scaled identically.  This keeps the chart from becoming lopsided toward the corpus that had a larger number of terms.\footnote{While both are available, ordinal ranks are preferable to log frequency since uninteresting stop-words often occupy disproportionate axis space.} 

\begin{figure}[h]
  \includegraphics[width=\linewidth]{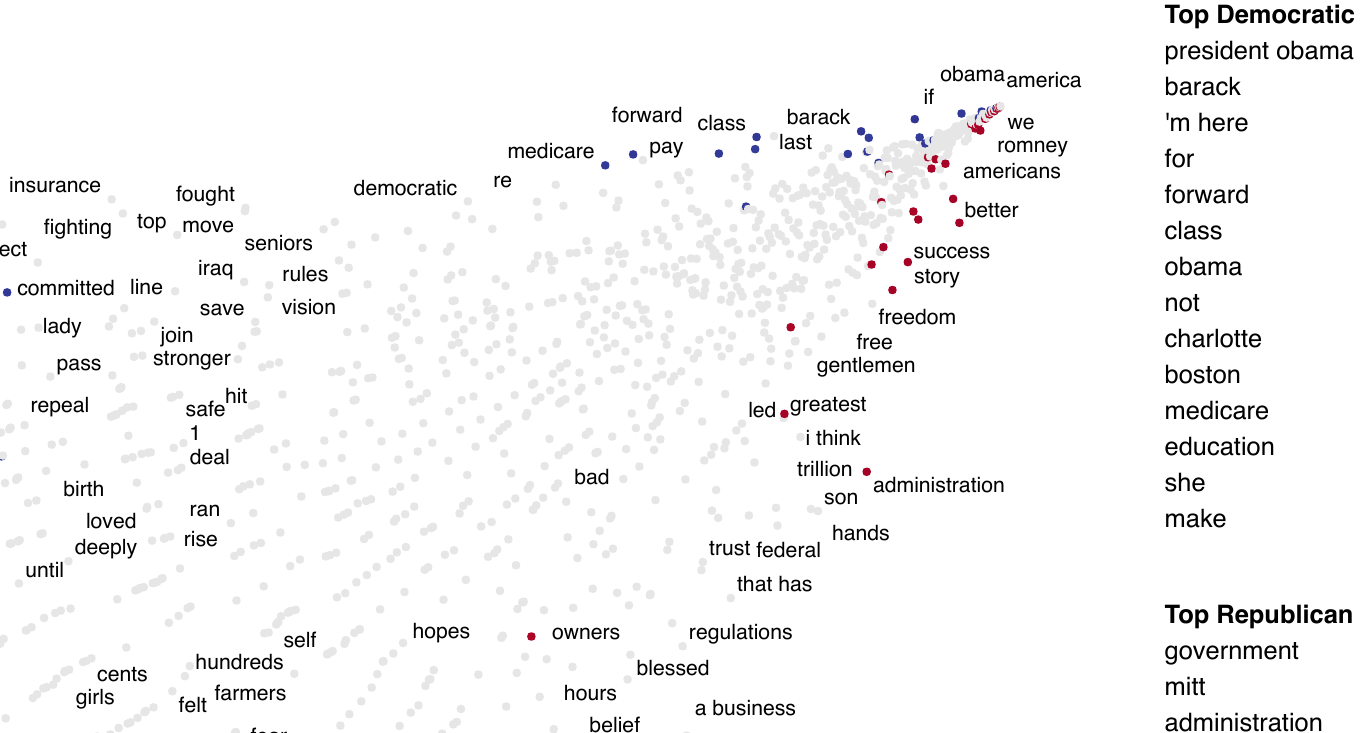}
  \caption{A cropped view of points being colored using $\ell$1-logreg coefficients. Interactive version: \href{https://jasonkessler.github.io/st-sparse.html}{jasonkessler.github.io/st-sparse.html}}
  \label{scattertextsparse}
\end{figure}

The charts in Figures \ref{scattertextmain}, \ref{scattertextsparse}, and \ref{scattertextembeddings}, were made with parameters $\mbox{\boldmath$m$}{=}5$, $\mbox{\boldmath$p$}{=}8$, and $\mbox{\boldmath$\phi$}{=}\phi^{T}$. 

\textbf{Breaking ties alphabetically is a simple but important alternative to jitter.} While jitter (i.e., randomly perturbing $x_{t}^{A}$ and $x_{t}^{B}$) breaks up the stacked points shown in Figure \ref{nojitterfig}, it eliminates empty space to legibly label points.  Jitter can make it seem like identically frequent points are closer to an upper left or lower right corner.  Alphabetic tie-breaking makes identical adjustments to both axes, leading to the horizontal (lower-left to upper-right) alignments of identically frequent points.  This angle does not cause one point to be substantially closer to either of the category associated corners (the upper-left and lower-right). 

These alignments provide two advantages. First, they open up point-free tracts in the center of the chart which allow for unobstructed interior labels. Second, they arrange points in a way that it is easy to hover a mouse over all of them, to indicate what term they correspond to, and be clicked to see excerpts of that term.

In the running example, 154 points were labeled when a jitter of 10\% of each axis and no tie-breaking was applied. 210 points (a 36\% lift) were labeled when no jitter was applied. 140 were labeled if no tie breaking was used.

Rudder \shortcite{Rudder2014} observed terms closer to the lower-right corner were used frequently in $A$ and infrequently in $B$, indicating they have both high recall and precision wrt category $A$.  Symmetrically, the same relationship exists for $B$ and the upper-right corner.  I can formalize this score between a point's coordinates and it's respective corner.  This intuition is represented by a score function $s_K(t)$ ($K\in\{A,B\}$ and $t\in V$) where
\vspace{-.3cm}
\begin{equation}
s_K(t)= 
\begin{cases} \|\langle 1-x_{t}^{A}, x_{t}^{B}\rangle\| & \text{if $K=A$,}
\\
\|\langle x_{t}^{A}, 1-x_{t}^{B}\rangle\| &\text{if $K=B$}
\end{cases}.
  \label{eqn:cornerscore}
\end{equation}
Other term scoring methods (e.g., regression weights or a weighted log-odd-ratio with a prior) may be used in place of Formula \ref{eqn:cornerscore}.

Maximal non-overlapping labeling of scatterplots is NP-hard \cite{Been2007}. Scattertext's heuristic is labeling points if space is available in one of many places around a point.  This is performed iteratively, beginning with points having the highest score (regardless of category) and proceeding downward in score.  An optimized data structure automatically constructed using Cozy \cite{cozy} holds the locations of drawn points and labels. 

The top scoring terms in classes $B$ and $A$ (Democrats and Republicans in Figure \ref{scattertextmain}) are listed to the right of the chart.  Hovering over points and terms highlights the point and displays frequency statistics.

Point colors are determined by their scores on $s$.  Those corresponding to terms with a high $s_B$ colored in progressively darker shades of blue, while those with a higher $s_A$ are colored in progressively darker shades of red.  When both scores are about equal, the point colors become more yellow, which creates a visual divide between the two classes.   The colors are provided by D3's ``RdYlBu'' diverging color scheme from Colorbrewer\footnote{\href{http://colorbrewer2.org/}{colorbrewer2.org}} via d3\footnote{\href{https://github.com/d3/d3-scale-chromatic}{github.com/d3/d3-scale-chromatic}}.

Other point colors (and scorings) can be used.  For example, Figure \ref{scattertextsparse} shows coefficients of an $\ell$1 penalized log. reg. classifier on $V$ features.  Scattertext, in this example, is set to color 0-scoring coefficients light gray.  Terms' univariate predictive power are still evident by their chart position.  See below\footnote{
\href{https://jasonkessler.github.io/sparseviz.html}{jasonkessler.github.io/sparseviz.html}} for an interactive version.
\section{Topical category discriminators}
\begin{figure}[h]
  \includegraphics[width=\linewidth]{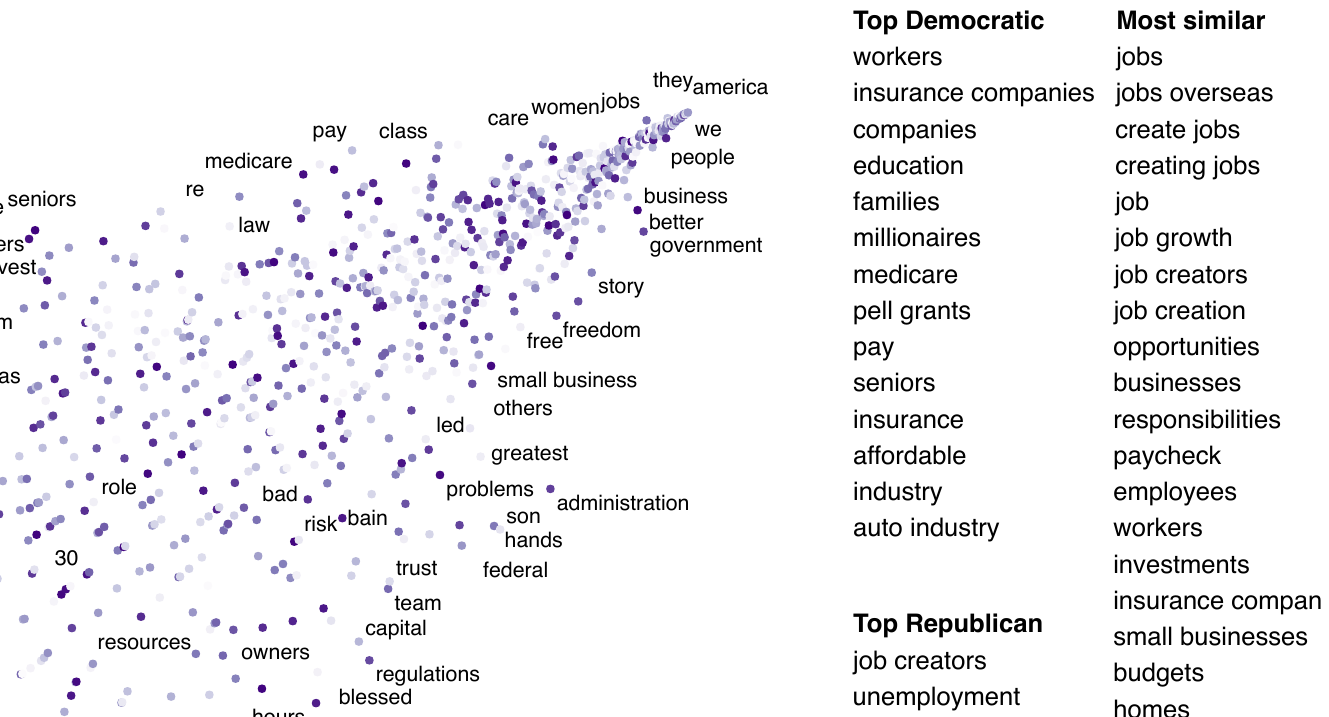}
  \caption{Words and phrases that are semantically similar to the word ``jobs'' are colored darker on a gray-to-purple scale, and general and category-specific related terms are listed to the right. Note that this is a cropping of the upper left-hand corner of the plot. Interactive version: \href{https://jasonkessler.github.io/st-sim.html}{jasonkessler.github.io/st-sim.html}.}
  \label{scattertextembeddings}
\end{figure}
\label{embeddings}
In 2012, how did Republicans and Democrats use language relating to ``jobs'', ``healthcare'', or ``military'' differently?   Figure \ref{scattertextembeddings} shows, in the running example, words similar to ``jobs'' that were characteristic of the political parties.  

In this configuration of Scattertext, words are colored by their cosine similarity to a query phrase.  This is done using spaCy\footnote{\href{https://spacy.io/}{spacy.io}}-provided GloVe \cite{glove} word vectors (trained on the Common Crawl corpus). Mean vectors are used for phrases.

The calculation of the most similar terms associated with each category is a simple heuristic.  First, sets of terms closely associated with a category are found. Second, these terms are ranked based on their similarity to the query, and the top rank terms are displayed to the right of the scatterplot (Figure \ref{scattertextembeddings}).  

A term is considered associated if its p-value is $<$0.05.  P-values are determined using MCQ's difference in the weighted log-odds-ratio with an uninformative Dirichlet prior.  This is the only model-based method discussed in Monroe et al. that does not rely on a large in-domain background corpus.  Since I am scoring bigrams in addition to the unigrams scored by MCQ, the size of the corpus would have to be larger to have high enough bigram counts for proper penalization.

This function relies the Dirichlet distribution's parameter $\alpha \in \mathbf{R}_{+}^{|V|}$. Following MCQ, $\alpha_t = 0.01$.  Formulas 16, 18 and 22 are used to compute z-scores, which are then converted to p-values using the Normal CDF of $\hat{\zeta}_{w}^{A-B}$, letting $y^{(K)}_{t} = \mbox{\boldmath$\phi$}(t,K)$ st ${K}\in{\{A,B\}}$ and $t\in{V}$.

As seen in Figure \ref{scattertextembeddings}, the top Republican word related to ``jobs'' is ``job creators'', while ``workers'' is the top Democratic term.
\section{Conclusion and future work}
Scattertext, a tool to make legible, comprehensive visualizations of class-associated term frequencies, was introduced.  Future work will involve rigorous human evaluation of the usefulness of the visualization strategies discussed.
\section*{Acknowledgments}
Jay Powell, Kyle Lo, Ray Little-Herrick, Will Headden, Chuck Little, Nancy Kessler and Kam Woods helped proofread this work.
\vspace{-.5cm}
\bibliography{kessler2017}
\bibliographystyle{acl_natbib}
\end{document}